\documentclass[sigconf, nonacm]{acmart}

\AtBeginDocument{%
  }

\usepackage{listings}
\usepackage{xcolor}

\usepackage{verbatim}

\usepackage{xspace}
\xspaceaddexceptions{]\}}

\usepackage{hyperref}
\usepackage{tabularx}
\usepackage{makecell}
\usepackage{colortbl}
\usepackage{ltablex}
\usepackage{tablefootnote}

\usepackage{enumitem}

\newlist{questions}{enumerate}{2}
\setlist[questions,1]{label=RQ\arabic*:,ref=RQ\arabic*}

\definecolor{darkgreen}{rgb}{0.05,0.5,0.05}
	\definecolor{denim}{rgb}{0.08, 0.38, 0.74}
\definecolor{darkviolet}{rgb}{0.58, 0.0, 0.83}

\begin{document}

\title{IRL Dittos: Embodied Multimodal AI Agent Interactions in Open Spaces}

\author{Seonghee Lee}
\email{shl1027@stanford.edu}
\affiliation{%
  \institution{Stanford University}
  \city{Stanford}
  \state{California}
  \country{USA}
}

\author{Denae Ford}
\email{denae@microsoft.com}
\affiliation{%
  \institution{Microsoft Research}
  \city{Redmond}
  \state{Washington}
  \country{USA}
}

\author{John Tang}
\email{johntang@microsoft.com}
\affiliation{%
  \institution{Microsoft Research}
  \city{Redmond}
  \state{Washington}
  \country{USA}
}
\author{Sasa Junuzovic}
\email{sasajun@microsoft.com}
\affiliation{%
  \institution{Microsoft Research}
  \city{Redmond}
  \state{Washington}
  \country{USA}
}
\author{Asta Roseway}
\email{astar@microsoft.com}
\affiliation{%
  \institution{Microsoft Research}
  \city{Redmond}
  \state{Washington}
  \country{USA}
}

\author{Ed Cutrell}
\email{cutrell@microsoft.com}
\affiliation{%
  \institution{Microsoft Research}
  \city{Redmond}
  \state{Washington}
  \country{USA}
}

\author{Kori Inkpen}
\email{kori@microsoft.com}
\affiliation{%
  \institution{Microsoft Research}
  \city{Redmond}
  \state{Washington}
  \country{USA}
}

\renewcommand{\shortauthors}{Lee et al.}

\begin{abstract}
We introduce the In Real Life (IRL) Ditto—an AI-driven, embodied agent designed to represent remote colleagues in shared office spaces, creating opportunities for real-time exchanges even in their absence. IRL Ditto offers a unique hybrid experience by allowing in-person colleagues to encounter a digital version of their remote teammates, initiating greetings, updates, or small talk as they might in person. Our research question examines: How can the IRL Ditto influence interactions and different relationships among colleagues in a shared office space? Through a four-day study, we assessed IRL Ditto’s ability to strengthen social ties by simulating presence and enabling meaningful interactions across different levels of social familiarity. We find that enhancing social relationships relied deeply on the foundation of the relationship participants had with the Source of the IRL Ditto. This study provides insights into the role of embodied agents in enriching workplace dynamics for distributed teams.
\end{abstract}

\begin{teaserfigure}
  \includegraphics[width=\textwidth]{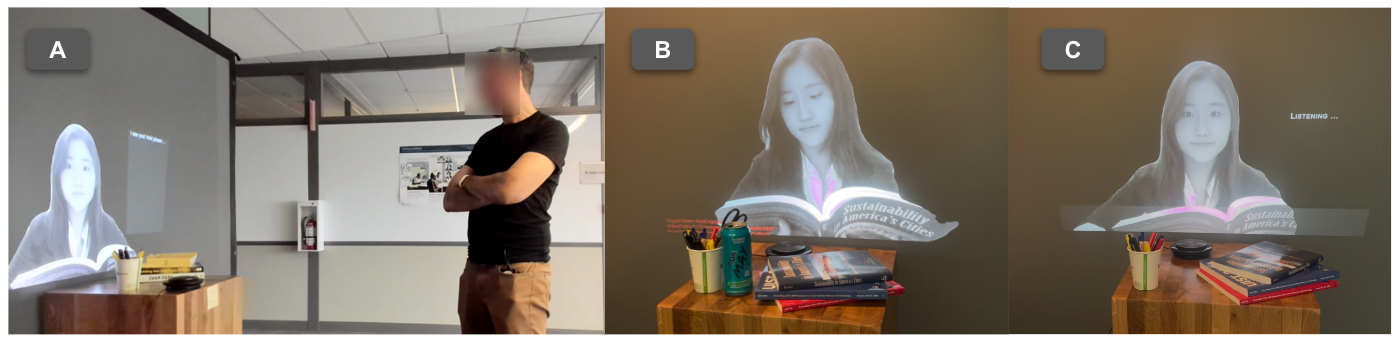}
  \caption{\textbf{Interaction Scene of In Real Life (IRL) Dittos} 
  (A) The IRL Ditto projected onto an office hallway with an engaging participant. 
  (B) The default state of the IRL Ditto - reading a book with motions of flipping a page. 
  (C) The IRL Ditto responding to a passerby displaying interaction intent.}
  \Description{}
  \label{fig:teaser}
\end{teaserfigure}

\maketitle

\section{Introduction}

The In Real Life (IRL) Ditto is an AI-driven, embodied autonomous agent designed to represent remote colleagues in shared office spaces, creating opportunities for impromptu, real-time exchanges even when the person is physically absent. The term Source, refers to the person that would be represented by the IRL Ditto \cite{leong2024dittos}. The IRL Ditto provides an opportunity for the Source to participate in spontaneous hallway interactions that otherwise would have been unavailable if the Source were remote. It initiates context-appropriate exchanges, such as greetings, updates, or brief conversations, based on proximity of the passersby. We explore the role of the IRL Ditto in fostering social connection and engagement on the job, focusing on the research question \emph{How can the IRL Ditto influence interactions and relationships among colleagues in a shared office space?}. To investigate this, we conduct a study in which the IRL Ditto is placed in an office hallway (see Figure~\ref{fig:teaser}) and engages with participants with social relationships to the Source as well as those with prior relationships. By comparing these interactions of people with different social relations to the Source, we aim to understand how personalized engagement differs from generic conversations and how these dynamics affect the experience of social connection. Our analysis explores how the IRL Ditto can go beyond simulating presence to actively enrich workplace dynamics. By leveraging proxemic cues and personalized knowledge of individuals with prior social relationships with the Source, we examine the IRL Ditto's ability to respond in contextually appropriate ways and foster social connections. We explore some challenges it faces on conversational dynamics, such as turn-taking and responsiveness or needs for social repair. By situating the IRL Ditto in an office hallway, we aim to find opportunities for collaboration and inclusion, helping remote employees maintain strong social ties while uncovering the challenges of deploying socially and contextually aware AI-proxies in real time.

\section{Related Work}

\subsection{AI Proxies in Collaborative Workspaces}

The concept of AI proxies for remote participants has evolved to include agents that are digital embodiments of users, engaging in both formal and informal interactions. Recent work, such as AI Delegates and Digital Agents in Meetings (Dittos)~\cite{leong2024dittos, chen2024aidelegatesdualfocus}, demonstrate how autonomous agents can effectively represent users by mimicking their preferences, engagement style, and communication patterns. These agents handle task-focused interactions and facilitate casual, relationship-building exchanges, reinforcing their role as proxies that bridge the gap between presence and absence. Past work involving Dittos~\cite{leong2024dittos}, personalized, embodied agents that encapsulate a user’s knowledge, appearance, and behavior found that Dittos serve as extensions of the user, participating in meetings on their behalf while maintaining a sense of familiarity and continuity. However, such systems also introduce unique challenges, including the potential for agents to disrupt conversational dynamics, users' hesitancy to challenge AI decisions, and the risks of embedding or amplifying biases within these systems.
Building on these ideas, our work explores how AI proxies can extend beyond virtual meetings to seamlessly integrate into physical spaces.

\subsection{Fostering Social Connection through Life-Sized AI and Digital Representation}

Creating meaningful social connections between remote and in-person team members remains a persistent challenge in hybrid environments~\cite{perspectives}. Prior research has explored embodied conversational agents using life-sized projections and robotic representations to simulate presence and facilitate engagement~\cite{room2room, directionRobot}. While these methods effectively leverage shared spatiality to foster inclusion, they are inherently limited to ``same-time, different-place'' interactions, requiring synchronous engagement to enable collaboration. This constraint often excludes scenarios where participants cannot be present simultaneously, hindering equitable collaboration across time. Emerging AI technologies offer the potential to bridge this gap by dynamically generating or ``hallucinating'' context required to support asynchronous collaboration. Systems such as the Dimensions in Testimony exhibit~\cite{DimensionsInTestimony} demonstrate how AI-driven conversational interactions with pre-recorded individuals can create a sense of presence, enabling participants to engage as if in real-time. Although not fully dynamic, these examples highlight how AI can simulate temporal co-presence, fostering meaningful connections even when live participants are absent.
Recent advances extend these capabilities to professional and social contexts, such as service industries and Digital Twins~\cite{HoffmanAITwin, SamsungDigitalHumans}, which offer tailored, adaptive interactions. Additionally, the Honeypot Effect~\cite{honeypot} illustrates how a public installation in an open space can passively engage participants, drawing them into interaction without explicit initiation from the system or the participants themselves. By combining these insights, our study investigates how AI-powered life-sized representations in open spaces can foster deeper social connections by enabling both synchronous and asynchronous engagement, thus addressing the limitations of current systems.

\vspace{-.5em}
\subsection{Proxemic Interactions and Digital Agents in Open Spaces}

Research on proxemic interactions~\cite{proxemic, crossProxemic, digitalProxemic} highlights the importance of spatial distance and body orientation (proxemic cues) in designing intuitive and engaging interactions. Proxemic responsiveness—where agents adjust behavior based on proximity and movement—enhances perceived intelligence and social presence by mimicking human nonverbal cues such as gaze, gestures, and posture. Studies show that robot personality traits can influence human proxemic behavior, with individuals maintaining a shorter distance from robots with introverted traits compared to extroverted ones during verbal interactions~\cite{robotProxemics}. Comparative analyzes have also revealed differences in the preferences of proxemics between augmented reality interfaces and physically embodied robots~\cite{Nigro_2024}. However, limited information about proxemic perferences for digital projected agents such as the IRL Ditto is known. Moreover, systems like the Platform for Situated Intelligence (PSI) leverage multimodal data to enable real-time interactions that are crucial for proxemic systems, while F-formations explore spatial dynamics and body orientation in multi-user interactions~\cite{psi, cross-device, F-formations}. Building on this previous work, the IRL Ditto is designed to take advantage of proxemic cues, such as distance, body orientation, and familiarity with interaction partners, to tailor its behavior to specific interaction contexts and user expectations, fostering seamless human-agent collaboration.

\section{System Overview}
In this section, we highlight the overview of the system design of the IRL Ditto and explains the design of the interaction with the IRL Ditto.

\vspace{-.75em}
\subsection{Agent Representation}
The IRL Ditto is designed to facilitate dynamic, immersive interactions by combining synchronized lip synching, voice cloning, and personalized knowledge. It operates through a series of video streams that simulate various behaviors, such as being idle, listening, or greeting others. 
In this system, we refer to the user that this embodied agent represents as the Source~\cite{leong2024dittos}.
The agent's voice is a cloned version of the Source’s voice, created using Azure AI Services Speech technology. In terms of personality, the agent is configured using the GPT-4o large language model and was set to reflect traits provided by the Source, such as being slightly more outgoing. This adjustment takes into account the context in which the Source is placed-—specifically, an office hallway where the agent is intended to engage with the Source's colleagues. While the agent mimics the Source's appearance and voice, certain aspects, like the representation of the video or the delay in its responses, may help users recognize that this is not the Source themselves. These subtle differences in interaction contribute to a clear distinction between the agent and the original person.

\subsection{Setup and Devices}
The IRL Ditto is visually represented by a life-sized projected screen of the head and torso of the IRL Ditto, accompanied by a desk with books to enhance realism without rendering the full body. An Azure Kinect camera captures individuals' distance and orientation, while Ultra-Wideband (UWB) sensors identify specific individuals selected for the study through a nearby receiver.
The IRL Ditto's audio is rendered using Azure AI services and the Ditto's lip synchronization was rendered real-time on a desktop machine with an Nvidia 4090 GPU.

\subsection{Agent Interaction Transition Sequence}

\begin{figure*}[!t]
    \centering \includegraphics[width=1.0\textwidth]{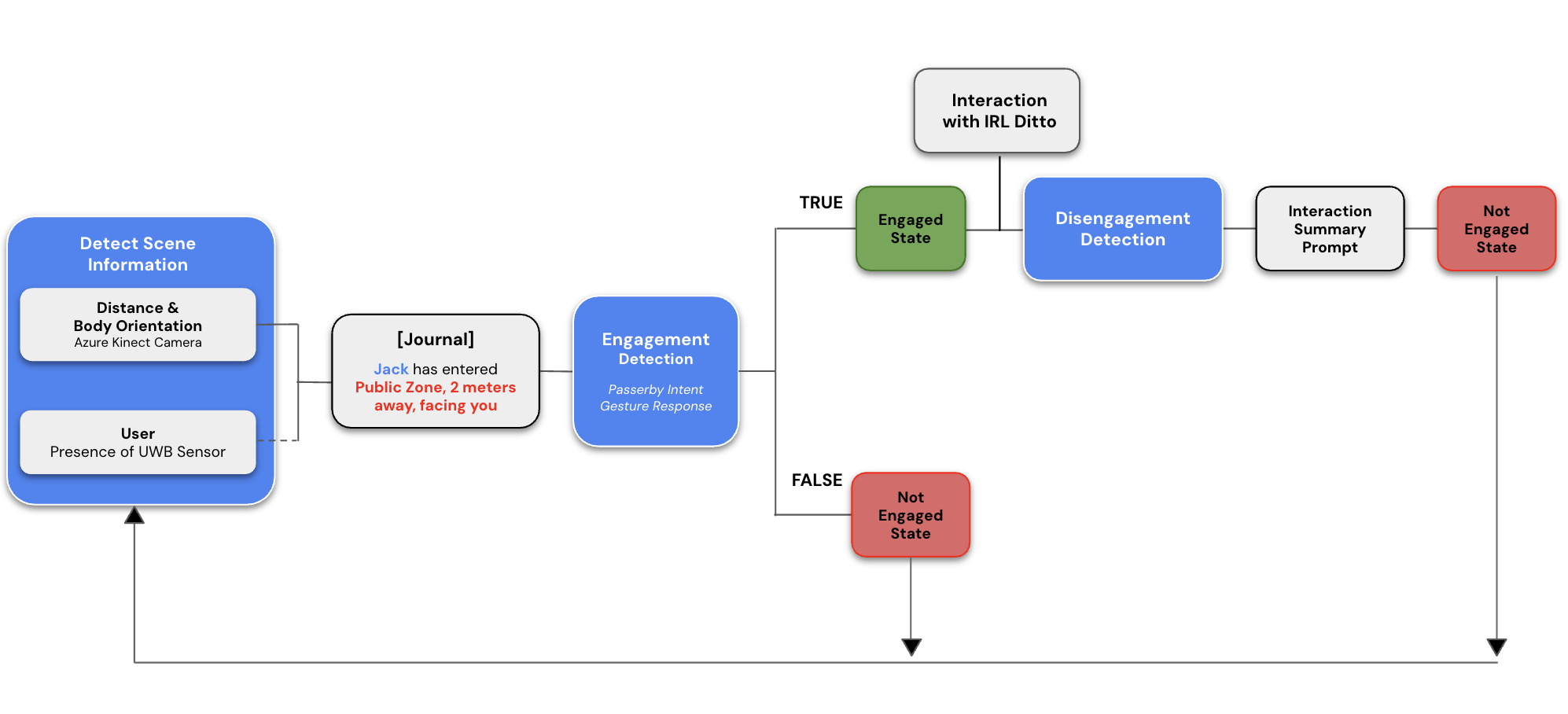}
    \caption[Flowchart of Interaction Mechanism in the IRL Ditto Zone]{
        At the start, the IRL Ditto begins in the \textit{"Not Engaged"} state, where a Kinect camera and UWB sensors detect individuals and log proximity and identification information within the journal. The model then makes an engagement check decision based on this logged information. If there is an interaction intent from the participant, the agent transitions to the \textit{"Engaged"} state. During engagement, the conversation continues, and periodic disengagement checks are performed. If disengagement is detected, the system returns to the \textit{"Not Engaged"} state.
    }
    \label{fig:flowchart}
\end{figure*}

\subsubsection{Detecting a Person in the Scene}
The agent begins in a Not Engaged state, appearing to read a book. When the Azure Kinect detects an individual entering the environment, it captures their distance and orientation, logging this in the JOURNAL as ``\textbf{Passerby} has entered the zone, 2 meters away, facing you.'' For recognized individuals with UWB sensors, the log includes their name, e.g., ``\textbf{Jack} has entered the public zone, 2 meters away, facing you.'' The agent then uses this spatial and identity data to assess engagement potential through the Engagement Detection System Prompt (Appendix A).

\subsubsection{Entering the Engaged State}
If interaction intent is detected—based on proxemic data (distance, orientation), inferred social relationship, or spoken cues—the agent transitions into the Engaged State. This shift from passive to active engagement occurs when a person moves closer, enters a social zone, or addresses the agent directly. The agent initiates conversation in the Source’s voice, with lip synchronization via Wav2Lip~\cite{prajwal2020wav2lip}, ensuring natural alignment with visual cues. The presence of a tag influences the personalization of interactions, with tagged participants receiving more tailored prompts, while untagged passersby experience more neutral responses. The presence of a tag influences the level of personalization, with tagged participants receiving prompts tailored to their pre-collected data, while untagged passersby experience general, context-agnostic responses.

\subsubsection{Engaging in the Conversation}
In the Engaged State, the agent responds in real-time to the user’s speech input. The conversation evolves based on the individual’s inferred intent and background information known to the agent. For instance, if the user asks a question, the agent provides contextually appropriate responses, and may also ask questions aligned with the Source’s intended topics, enhancing depth. Throughout, the agent’s visuals are designed to maintain eye contact and nod naturally, reinforcing the sense of personalized engagement.

\subsubsection{Leaving, Pausing, and Ending the Conversation} After several turns, the agent asks, ``Would you like to stay and chat a little longer?'' If the user chooses to leave, the agent politely ends the conversation and returns to its idle state. If the user walks away mid-conversation, it seamlessly returns to idle, mimicking casual hallway exchanges.

\subsubsection{Summarizing the Conversation}

At the end of each conversation, the system generates a brief summary capturing key points and highlights, such as discussed topics, decisions, or notable personal details for future interactions. Unique summaries are saved for tagged participants, enabling the agent to refer to prior interactions and maintain conversational continuity. For untagged Passerby participants, a general summary captures non-specific details to support broader contextual interactions. For Passerby participants, a general summary is maintained, enabling the agent to recall prior topics and interactions.

\section{Study Design and Procedure}

To address our research question, we conducted a four-day, IRB-approved study evaluating the IRL Ditto. Participants interacted with the agent daily, enabling us to observe its impact on social engagement.

\subsection{Setting: IRL Ditto Zone}
The IRL Ditto zone served as the dedicated interaction space, marked with white tape and signage to inform individuals that video recording was in progress. Signage indicated that entering the zone implied consent for recording, while an alternative route was provided for those who wished to avoid participation. This signage can be found in Figure~\ref{fig:study-signage}.
The zone was active for a designated two hour slot each day, capturing interactions from participants. Microsoft Teams was used to record interactions with the IRL Ditto. Nearby office workers were informed to prevent unintended audio capture, and Teams' noise suppression feature was utilized to minimize ambient sounds.

\subsection{Participants}
The study consisted of 3 types of participants: The Source, Tag-Wearing participants, and Passerby participants: 

\vspace{.5em}
\noindent\textbf{Source.}
The Source provided the agent’s visual and vocal representation through recorded videos and voice samples, creating a natural and lifelike interaction. The Source was an intern at the time of this study. The Source was briefed and consented to participate.

\vspace{.5em}
\noindent\textbf{Tag-Wearing Participants.} 
We recruited six participants, equipping them with UWB tags prior to the study to enable recognition within the zone. Each participant signed a consent form, and the Source provided relevant relationship descriptions and conversational topic information for personalized agent responses with participants. These participants completed a daily survey about their interactions and were interviewed after the study.

\vspace{.5em}
\noindent\textbf{Passerby Participants.} 
For passersby, consent was implied upon entry into the clearly marked IRL Ditto zone, as indicated by signage stating that stepping into the zone constituted consent. This signage was reviewed and approved by the Internal Review Board. Video recording was conducted through a web camera connected to Microsoft Teams, with noise suppression active. Passersby were invited to complete an optional survey after their interaction and select participants were asked if they could participate in an interview.

\vspace{.5em}

Upon conclusion of the study, 21 participants had interacted with the IRL Ditto. We conducted 13 post-study interviews including the Source, all six tag-wearing participants, and six passersby participants. All interview participants received remuneration in the form of a \$20 gift card. We then analyzed data from daily surveys and interview transcripts where we identified themes of experiences. Due to limited space, we highlight select findings from our analysis.

\begin{table*}[!h]
\centering
\begin{tabular}{|c|c|p{10cm}|}
\hline
\textbf{Participant} & \textbf{Participant Type} & \textbf{Social Relationship Description} \\ \hline
T1 & Tag-Wearing & Fellow intern friend of the Source, interacted in group settings only. \\ \hline
T2 & Tag-Wearing &  Fellow intern friend of the Source, who works on an adjacent team. Office is in closer proximity so sees more frequently. \\ \hline
T3 & Tag-Wearing & Fellow intern friend of the Source, attends same school, interacted in 1:1 settings as well as group settings.\\ \hline
T4 & Tag-Wearing & FTE on an adjacent team with the Source. Interacted in work meetings but never personally. \\ \hline
T5 & Tag-Wearing & FTE that works in the Hardware Lab, helped the Source on a lot of project setup. \\ \hline
T6 & Tag-Wearing & FTE that the Source interacted with during informal work settings and company poster sessions. \\ \hline
P1 & Passerby & FTE that has interacted with the source in both work settings and informal settings too.\\ \hline
P2 & Passerby & Close fellow intern that spent time with the Source on intern activities in informal settings.\\ \hline
P3 & Passerby & FTE with no prior interaction with the Source.\\ \hline
P4 & Passerby & FTE with no prior interaction with the Source.\\ \hline
P5 & Passerby &  FTE with no prior interaction with the Source. \\ \hline
P6 & Passerby & Fellow intern that spent time with the Source on office social activities in informal settings.\\ \hline
\end{tabular}
\caption{
Descriptions of the interview participants (12) of the study out of the total 21 participants that participated in the study. These participants were asked to approach the scene wearing a UWB sensor. A total of 12 participants were interviewed: 6 Tag-wearing and 6 Passersby.}
\end{table*}

\section{Results}
The study revealed a variety of interactions that the IRL Ditto helped create, showing how things like the participant’s relationship with the Source and the situation around them played a role in shaping their experiences. These results give us a closer look at how the IRL Ditto influenced social interactions, helping form connections, steer conversations, and deal with different challenges in real-time. Ultimately, the findings provide valuable insights into how the IRL Ditto interacts with people in everyday settings and adapts to different social contexts.

\subsection{Building Social Bonds and Fostering Relationships through the Ditto}

For participants who had an established relationship with the Source, the IRL Ditto’s presence often strengthened their connection. For example, T3 appreciated the IRL Ditto’s gestures, sharing, \textit{``I liked how the IRL Ditto initiated contact by waving; it felt natural, and asking about recent social media updates made the interaction feel familiar.''} Participants also leveraged the IRL Ditto to coordinate real-world connections, as T3 noted, \textit{``I made plans with the IRL Ditto to catch up with the Source at Coupa Cafe because I wanted to stay connected after our internship.''} 
While participants familiar with the Source appreciated the IRL Ditto’s ability to reference shared experiences or details, they occasionally found the IRL Ditto’s conversational liberties intrusive. For instance, T3 valued the IRL Ditto’s effort to correct a name mispronunciation but noted, \textit{``It felt a bit overdone and unnecessary.''} Similarly, P6, a friend of the Source, found certain corrections awkward, suggesting that the IRL Ditto’s efforts to manage social repair need to be more subtle.

\subsection{Interactions with the IRL Ditto Among Participants Unacquainted with the Source}

Participants unfamiliar with the Source often questioned the IRL Ditto’s identity and intent. T2 remarked, \textit{I wasn’t sure if I should treat the IRL Ditto as the Source or as an AI Agent—it was a bit unclear.''} T4 noted, \textit{It felt like it wasn’t entirely the Source, which made me hesitate.''} These comments underscore the need for the IRL Ditto to clarify its connection to the Source. Passersby also reported challenges engaging with the IRL Ditto. P3 shared, \textit{I didn’t feel like it was speaking to me directly—it was more like I was observing someone else’s interaction.''} P5 stated, \textit{It was tough to know when it was okay to speak—even with the onscreen cue.''} Despite these challenges, both P3 and P5 emphasized that their perception of the Source was not negatively impacted. Instead, they viewed these difficulties as limitations of the IRL Ditto itself, rather than a reflection of the Source. This distinction highlights the resilience of the Source’s identity, even when the IRL Ditto’s interactions faltered. However, some participants felt otherwise. T4 highlighted a deeper issue, questioning whether the IRL Ditto’s behavior reflected the Source: \textit{``Can I blame the IRL Ditto for being rude, or should I attribute these characteristics to the Source? If this duo were a real person, I wouldn’t interact with them. I wish I knew the Source better to form an accurate representation.''} This emphasizes the importance of seamless interactions and needs for social repair that preserve the Source’s identity and foster trust.

\subsection{Navigating Model Limitations and Social Repair Needs}

Timing and turn-taking were challenging, as the IRL Ditto often misinterpreted pauses as the end of a response and interrupted participants. T4 remarked, \textit{``It felt like [the] IRL Ditto wasn’t listening to what I was trying to say.''} This issue was particularly pronounced for participants like P2, who identified as having a speech impairment. Additionally, instances where the IRL Ditto provided incorrect information—such as mentioning a non-existent office birthday party—or inappropriately corrected the pronunciation of someone's name led to confusion and diminished user trust. We found it more effective for the IRL Ditto to ask users if they wanted to continue the conversation, as it often prematurely ended interactions based on its contextual judgment of a short hallway exchange. Additionally, T5 noted that the IRL Ditto mentioned their interest in ships, which the Source had learned indirectly. This unexpected reference surprised T5 and highlighted the importance of maintaining shared conversational grounding to ensure comfort and clarity in interactions.

\vspace{-1em}
\section{Discussion}

This study sheds light on both the potential and limitations of AI-driven social proxies like the IRL Ditto in fostering connection, navigating conversational missteps, and preserving the essence of the individual they represent. 

\vspace{-.5em}
\subsection{Social Repair Needs in the Interaction Process}

Effective conversational repair was crucial for maintaining meaningful interactions between participants and the IRL Ditto while supporting social relationships with the Source. Challenges arose when the IRL Ditto misinterpreted pauses as the end of a turn, prematurely responding and disrupting the interaction flow. Delays in responses compounded these issues, leading to disjointed conversations where outdated topics were addressed after users had moved on. These disruptions, along with instances of inaccurate or irrelevant information—such as mentioning non-existent events or mispronouncing names—diminished trust and the natural flow of interactions.
To address these challenges, the IRL Ditto needed to better detect and adapt to conversational cues, such as pauses, hesitation, or sentiment changes. For example, it could rephrase or clarify statements to resolve confusion or alert the Source to review and correct potentially problematic interactions or inaccurate information. Allowing participants to control conversation length, such as by having the Ditto ask if they wanted to continue talking, further improved flexibility and reduced premature endings. Improved conversational repair mechanisms like this could enhance the IRL Ditto's ability to bridge AI limitations and user expectations, enabling more seamless interactions.

\subsection{Expanding Use Cases and Future Iterations for the IRL Ditto}
Participants suggested introducing visual markers, such as watermarks, to clearly identify the IRL Ditto as an extension of the Source, helping users, especially those less familiar with the Source, understand its role as an automated proxy. Furthermore, the IRL Ditto's potential as both a conversational tool and an assistant for managing interactions suggests the need for adaptable boundaries that seamlessly shift between casual and formal interactions, fostering trust in its ability to serve diverse needs. Participants also saw the IRL Ditto’s potential as a personal confidant, providing reminders in the voice of a trusted colleague, assisting with scheduling availability, or clarifying low-priority tasks during focus periods. In addition, the IRL Ditto’s role could extend to allowing remote colleagues participate in office hallway, lunch, and water-cooler conversations. Together, these adaptations would make the IRL Dittos a more contextually aware, trusted, and flexible tool in various interaction scenarios. At the time of developing this prototype, the available technology imposed certain limitations. Real-time processing, such as constant streaming and updating of conversational data, was not yet feasible due to the lack of accessible APIs for large language models (LLMs). This restricted the IRL Ditto’s ability to handle conversational nuances effectively, such as managing silences, stopping when a user began speaking, or dynamically adjusting its responses. As technology evolves, newer systems could provide better responsiveness and better adaptability to conversation.

\section{Conclusion}

This study explored the role of the IRL Ditto, an embodied autonomous agent in fostering social relationships and engaging in spontaneous conversations in an office hallway. Our findings suggest that while the IRL Ditto has the potential to reinforce social bonds through personalized interactions, challenges remain in areas like turn-taking, contextual awareness, and responsiveness. Addressing these challenges by incorporating adaptive conversational repair strategies could enhance the effectiveness of AI social proxies. Additionally, we envision an expanded role for the IRL Ditto, not only as a conversational partner but also as an active intermediary capable of imagining interactions and reporting back to the Source. This capability could involve the IRL Ditto summarizing key moments, relaying essential updates, or flagging issues that require attention, thus enhancing its utility as a trusted proxy in professional and social contexts. Future iterations of the IRL Ditto could enhance its ability to foster seamless, meaningful, and inclusive interactions in diverse scenarios.

\bibliographystyle{ACM-Reference-Format}
\bibliography{ref}


\begin{thebibliography}{18}


\ifx \showCODEN    \undefined \def \showCODEN     #1{\unskip}     \fi
\ifx \showDOI      \undefined \def \showDOI       #1{#1}\fi
\ifx \showISBNx    \undefined \def \showISBNx     #1{\unskip}     \fi
\ifx \showISBNxiii \undefined \def \showISBNxiii  #1{\unskip}     \fi
\ifx \showISSN     \undefined \def \showISSN      #1{\unskip}     \fi
\ifx \showLCCN     \undefined \def \showLCCN      #1{\unskip}     \fi
\ifx \shownote     \undefined \def \shownote      #1{#1}          \fi
\ifx \showarticletitle \undefined \def \showarticletitle #1{#1}   \fi
\ifx \showURL      \undefined \def \showURL       {\relax}        \fi
\providecommand\bibfield[2]{#2}
\providecommand\bibinfo[2]{#2}
\providecommand\natexlab[1]{#1}
\providecommand\showeprint[2][]{arXiv:#2}

\bibitem[Andrist et~al\mbox{.}(2023)]%
        {psi}
\bibfield{author}{\bibinfo{person}{Sean Andrist}, \bibinfo{person}{Dan Bohus}, \bibinfo{person}{Zongjian Li}, {and} \bibinfo{person}{Mohammad Soleymani}.} \bibinfo{year}{2023}\natexlab{}.
\newblock \showarticletitle{Platform for Situated Intelligence and OpenSense: A Tutorial on Building Multimodal Interactive Applications for Research}. In \bibinfo{booktitle}{\emph{Companion Publication of the 25th International Conference on Multimodal Interaction}} (Paris, France) \emph{(\bibinfo{series}{ICMI '23 Companion})}. \bibinfo{publisher}{Association for Computing Machinery}, \bibinfo{address}{New York, NY, USA}, \bibinfo{pages}{105–106}.
\newblock
\showISBNx{9798400703218}
\urldef\tempurl%
\url{https://doi.org/10.1145/3610661.3617603}
\showDOI{\tempurl}


\bibitem[Ballendat et~al\mbox{.}(2010)]%
        {proxemic}
\bibfield{author}{\bibinfo{person}{Till Ballendat}, \bibinfo{person}{Nicolai Marquardt}, {and} \bibinfo{person}{Saul Greenberg}.} \bibinfo{year}{2010}\natexlab{}.
\newblock \showarticletitle{Proxemic interaction: designing for a proximity and orientation-aware environment}. In \bibinfo{booktitle}{\emph{ACM International Conference on Interactive Tabletops and Surfaces}} (Saarbr\"{u}cken, Germany) \emph{(\bibinfo{series}{ITS '10})}. \bibinfo{publisher}{Association for Computing Machinery}, \bibinfo{address}{New York, NY, USA}, \bibinfo{pages}{121–130}.
\newblock
\showISBNx{9781450303996}
\urldef\tempurl%
\url{https://doi.org/10.1145/1936652.1936676}
\showDOI{\tempurl}


\bibitem[Bohus et~al\mbox{.}(2014)]%
        {directionRobot}
\bibfield{author}{\bibinfo{person}{Dan Bohus}, \bibinfo{person}{C.W. Saw}, {and} \bibinfo{person}{Eric Horvitz}.} \bibinfo{year}{2014}\natexlab{}.
\newblock \showarticletitle{Directions robot: In-the-wild experiences and lessons learned}.
\newblock \bibinfo{journal}{\emph{13th International Conference on Autonomous Agents and Multiagent Systems, AAMAS 2014}}  \bibinfo{volume}{1} (\bibinfo{date}{01} \bibinfo{year}{2014}), \bibinfo{pages}{637--644}.
\newblock


\bibitem[Chen et~al\mbox{.}(2024)]%
        {chen2024aidelegatesdualfocus}
\bibfield{author}{\bibinfo{person}{Xi Chen}, \bibinfo{person}{Zhiyang Zhang}, \bibinfo{person}{Fangkai Yang}, \bibinfo{person}{Xiaoting Qin}, \bibinfo{person}{Chao Du}, \bibinfo{person}{Xi Cheng}, \bibinfo{person}{Hangxin Liu}, \bibinfo{person}{Qingwei Lin}, \bibinfo{person}{Saravan Rajmohan}, \bibinfo{person}{Dongmei Zhang}, {and} \bibinfo{person}{Qi Zhang}.} \bibinfo{year}{2024}\natexlab{}.
\newblock \bibinfo{title}{AI Delegates with a Dual Focus: Ensuring Privacy and Strategic Self-Disclosure}.
\newblock
\newblock
\showeprint[arxiv]{2409.17642}~[cs.AI]
\urldef\tempurl%
\url{https://arxiv.org/abs/2409.17642}
\showURL{%
\tempurl}


\bibitem[Hedayati et~al\mbox{.}(2020)]%
        {F-formations}
\bibfield{author}{\bibinfo{person}{Hooman Hedayati}, \bibinfo{person}{Daniel Szafir}, {and} \bibinfo{person}{Sean Andrist}.} \bibinfo{year}{2020}\natexlab{}.
\newblock \showarticletitle{Recognizing F-formations in the open world}. In \bibinfo{booktitle}{\emph{Proceedings of the 14th ACM/IEEE International Conference on Human-Robot Interaction}} (Daegu, Republic of Korea) \emph{(\bibinfo{series}{HRI '19})}. \bibinfo{publisher}{IEEE Press}, \bibinfo{pages}{558–559}.
\newblock
\showISBNx{9781538685556}


\bibitem[Leong et~al\mbox{.}(2024)]%
        {leong2024dittos}
\bibfield{author}{\bibinfo{person}{Joanne Leong}, \bibinfo{person}{Ed Cutrell}, \bibinfo{person}{John Tang}, \bibinfo{person}{Sasa Junuzovic}, \bibinfo{person}{Gregory~Paul Baribault}, {and} \bibinfo{person}{Kori Inkpen}.} \bibinfo{year}{2024}\natexlab{}.
\newblock \showarticletitle{Dittos: Personalized, Embodied Agents That Participate in Meetings When You Are Unavailable}. In \bibinfo{booktitle}{\emph{2024 Conference on Computer Supported Cooperative Work}}. ACM, \bibinfo{publisher}{ACM}, \bibinfo{pages}{1--28}.
\newblock
\urldef\tempurl%
\url{https://www.microsoft.com/en-us/research/publication/dittos-personalized-embodied-agents-that-participate-in-meetings-when-you-are-unavailable/}
\showURL{%
\tempurl}


\bibitem[Marquardt et~al\mbox{.}(2012a)]%
        {crossProxemic}
\bibfield{author}{\bibinfo{person}{Nicolai Marquardt}, \bibinfo{person}{Ken Hinckley}, {and} \bibinfo{person}{Saul Greenberg}.} \bibinfo{year}{2012}\natexlab{a}.
\newblock \showarticletitle{Cross-device interaction via micro-mobility and f-formations}. In \bibinfo{booktitle}{\emph{Proceedings of the 25th Annual ACM Symposium on User Interface Software and Technology}} (Cambridge, Massachusetts, USA) \emph{(\bibinfo{series}{UIST '12})}. \bibinfo{publisher}{Association for Computing Machinery}, \bibinfo{address}{New York, NY, USA}, \bibinfo{pages}{13–22}.
\newblock
\showISBNx{9781450315807}
\urldef\tempurl%
\url{https://doi.org/10.1145/2380116.2380121}
\showDOI{\tempurl}


\bibitem[Marquardt et~al\mbox{.}(2012b)]%
        {cross-device}
\bibfield{author}{\bibinfo{person}{Nicolai Marquardt}, \bibinfo{person}{Ken Hinckley}, {and} \bibinfo{person}{Saul Greenberg}.} \bibinfo{year}{2012}\natexlab{b}.
\newblock \showarticletitle{Cross-device interaction via micro-mobility and f-formations}. In \bibinfo{booktitle}{\emph{Proceedings of the 25th Annual ACM Symposium on User Interface Software and Technology}} (Cambridge, Massachusetts, USA) \emph{(\bibinfo{series}{UIST '12})}. \bibinfo{publisher}{Association for Computing Machinery}, \bibinfo{address}{New York, NY, USA}, \bibinfo{pages}{13–22}.
\newblock
\showISBNx{9781450315807}
\urldef\tempurl%
\url{https://doi.org/10.1145/2380116.2380121}
\showDOI{\tempurl}


\bibitem[Moujahid et~al\mbox{.}(2023)]%
        {robotProxemics}
\bibfield{author}{\bibinfo{person}{Meriam Moujahid}, \bibinfo{person}{David~A. Robb}, \bibinfo{person}{Christian Dondrup}, {and} \bibinfo{person}{Helen Hastie}.} \bibinfo{year}{2023}\natexlab{}.
\newblock \bibinfo{title}{Come Closer: The Effects of Robot Personality on Human Proxemics Behaviours}.
\newblock
\newblock
\showeprint[arxiv]{2309.02979}~[cs.RO]
\urldef\tempurl%
\url{https://arxiv.org/abs/2309.02979}
\showURL{%
\tempurl}


\bibitem[Nigro et~al\mbox{.}(2024)]%
        {Nigro_2024}
\bibfield{author}{\bibinfo{person}{Massimiliano Nigro}, \bibinfo{person}{Amy O’Connell}, \bibinfo{person}{Thomas Groechel}, \bibinfo{person}{Anna-Maria Velentza}, {and} \bibinfo{person}{Maja Matarić}.} \bibinfo{year}{2024}\natexlab{}.
\newblock \showarticletitle{An Interactive Augmented Reality Interface for Personalized Proxemics Modeling: Comfort and Human–Robot Interactions}.
\newblock \bibinfo{journal}{\emph{IEEE Robotics\& Automation Magazine}} (\bibinfo{year}{2024}), \bibinfo{pages}{2–11}.
\newblock
\showISSN{1558-223X}
\urldef\tempurl%
\url{https://doi.org/10.1109/mra.2024.3415108}
\showDOI{\tempurl}


\bibitem[Pejsa et~al\mbox{.}(2016)]%
        {room2room}
\bibfield{author}{\bibinfo{person}{Tomislav Pejsa}, \bibinfo{person}{Julian Kantor}, \bibinfo{person}{Hrvoje Benko}, \bibinfo{person}{Eyal Ofek}, {and} \bibinfo{person}{Andrew Wilson}.} \bibinfo{year}{2016}\natexlab{}.
\newblock \showarticletitle{Room2Room: Enabling Life-Size Telepresence in a Projected Augmented Reality Environment}. In \bibinfo{booktitle}{\emph{Proceedings of the 19th ACM Conference on Computer-Supported Cooperative Work \& Social Computing}} (San Francisco, California, USA) \emph{(\bibinfo{series}{CSCW '16})}. \bibinfo{publisher}{Association for Computing Machinery}, \bibinfo{address}{New York, NY, USA}, \bibinfo{pages}{1716–1725}.
\newblock
\showISBNx{9781450335928}
\urldef\tempurl%
\url{https://doi.org/10.1145/2818048.2819965}
\showDOI{\tempurl}


\bibitem[Prajwal et~al\mbox{.}(2020)]%
        {prajwal2020wav2lip}
\bibfield{author}{\bibinfo{person}{K~R Prajwal}, \bibinfo{person}{Rudrabha Mukhopadhyay}, \bibinfo{person}{Vinay~P. Namboodiri}, {and} \bibinfo{person}{C.V. Jawahar}.} \bibinfo{year}{2020}\natexlab{}.
\newblock \showarticletitle{A Lip Sync Expert Is All You Need for Speech to Lip Generation In the Wild}. In \bibinfo{booktitle}{\emph{Proceedings of the 28th ACM International Conference on Multimedia}} (Seattle, WA, USA) \emph{(\bibinfo{series}{MM '20})}. \bibinfo{publisher}{Association for Computing Machinery}, \bibinfo{address}{New York, NY, USA}, \bibinfo{pages}{484–492}.
\newblock
\showISBNx{9781450379885}
\urldef\tempurl%
\url{https://doi.org/10.1145/3394171.3413532}
\showDOI{\tempurl}


\bibitem[{Reid Hoffman}(nd)]%
        {HoffmanAITwin}
\bibfield{author}{\bibinfo{person}{{Reid Hoffman}}.} \bibinfo{year}{n.d.}\natexlab{}.
\newblock \bibinfo{title}{Reid Hoffman meets his AI twin - Full}.
\newblock
\newblock
\urldef\tempurl%
\url{https://www.youtube.com/watch?v=rgD2gmwCS10}
\showURL{%
\tempurl}
\newblock
\shownote{Accessed: 2024-12-02}.


\bibitem[{Samsung Research}(nd)]%
        {SamsungDigitalHumans}
\bibfield{author}{\bibinfo{person}{{Samsung Research}}.} \bibinfo{year}{n.d.}\natexlab{}.
\newblock \bibinfo{title}{Digital Humans Research}.
\newblock
\newblock
\urldef\tempurl%
\url{https://sra.samsung.com/research-area/digital-humans/}
\showURL{%
\tempurl}
\newblock
\shownote{Accessed: 2024-12-02}.


\bibitem[Tang et~al\mbox{.}(2023)]%
        {perspectives}
\bibfield{author}{\bibinfo{person}{John~C. Tang}, \bibinfo{person}{Kori Inkpen}, \bibinfo{person}{Sasa Junuzovic}, \bibinfo{person}{Keri Mallari}, \bibinfo{person}{Andrew~D. Wilson}, \bibinfo{person}{Sean Rintel}, \bibinfo{person}{Shiraz Cupala}, \bibinfo{person}{Tony Carbary}, \bibinfo{person}{Abigail Sellen}, {and} \bibinfo{person}{William~A.S. Buxton}.} \bibinfo{year}{2023}\natexlab{}.
\newblock \showarticletitle{Perspectives: Creating Inclusive and Equitable Hybrid Meeting Experiences}.
\newblock \bibinfo{journal}{\emph{Proc. ACM Hum.-Comput. Interact.}} \bibinfo{volume}{7}, \bibinfo{number}{CSCW2}, Article \bibinfo{articleno}{351} (\bibinfo{date}{Oct.} \bibinfo{year}{2023}), \bibinfo{numpages}{25}~pages.
\newblock
\urldef\tempurl%
\url{https://doi.org/10.1145/3610200}
\showDOI{\tempurl}


\bibitem[{The Nancy \& David Wolf Holocaust \& Humanity Center}(nd)]%
        {DimensionsInTestimony}
\bibfield{author}{\bibinfo{person}{{The Nancy \& David Wolf Holocaust \& Humanity Center}}.} \bibinfo{year}{n.d.}\natexlab{}.
\newblock \bibinfo{title}{Dimensions in Testimony Exhibit}.
\newblock
\newblock
\urldef\tempurl%
\url{https://www.holocaustandhumanity.org/exhibits/dimensions-in-testimony/}
\showURL{%
\tempurl}
\newblock
\shownote{Accessed: 2024-12-02}.


\bibitem[Williamson et~al\mbox{.}(2022)]%
        {digitalProxemic}
\bibfield{author}{\bibinfo{person}{Julie~R. Williamson}, \bibinfo{person}{Joseph O'Hagan}, \bibinfo{person}{John~Alexis Guerra-Gomez}, \bibinfo{person}{John~H Williamson}, \bibinfo{person}{Pablo Cesar}, {and} \bibinfo{person}{David~A. Shamma}.} \bibinfo{year}{2022}\natexlab{}.
\newblock \showarticletitle{Digital Proxemics: Designing Social and Collaborative Interaction in Virtual Environments}. In \bibinfo{booktitle}{\emph{Proceedings of the 2022 CHI Conference on Human Factors in Computing Systems}} (New Orleans, LA, USA) \emph{(\bibinfo{series}{CHI '22})}. \bibinfo{publisher}{Association for Computing Machinery}, \bibinfo{address}{New York, NY, USA}, Article \bibinfo{articleno}{423}, \bibinfo{numpages}{12}~pages.
\newblock
\showISBNx{9781450391573}
\urldef\tempurl%
\url{https://doi.org/10.1145/3491102.3517594}
\showDOI{\tempurl}


\bibitem[Wouters et~al\mbox{.}(2016)]%
        {honeypot}
\bibfield{author}{\bibinfo{person}{Niels Wouters}, \bibinfo{person}{John Downs}, \bibinfo{person}{Mitchell Harrop}, \bibinfo{person}{Travis Cox}, \bibinfo{person}{Eduardo Oliveira}, \bibinfo{person}{Sarah Webber}, \bibinfo{person}{Frank Vetere}, {and} \bibinfo{person}{Andrew Vande~Moere}.} \bibinfo{year}{2016}\natexlab{}.
\newblock \showarticletitle{Uncovering the Honeypot Effect: How Audiences Engage with Public Interactive Systems}. In \bibinfo{booktitle}{\emph{Proceedings of the 2016 ACM Conference on Designing Interactive Systems}} (Brisbane, QLD, Australia) \emph{(\bibinfo{series}{DIS '16})}. \bibinfo{publisher}{Association for Computing Machinery}, \bibinfo{address}{New York, NY, USA}, \bibinfo{pages}{5–16}.
\newblock
\showISBNx{9781450340311}
\urldef\tempurl%
\url{https://doi.org/10.1145/2901790.2901796}
\showDOI{\tempurl}


\end{thebibliography}

\section{Appendix}

\appendix

\section{Engagement Detection Prompt}

\begin{lstlisting}
Choose one of the topics of the day to talk about. You should choose different topic that is not in the summary. 

Topic of the Day: 
Topic 1 Pottery: Today you have a nice picture of the pottery you made in front of you. You made this after work yesterday in the pottery class event. You want to ask people about what they think of your work. It's the last event before you leave!


\end{lstlisting}

\section{Disengagement Detection Prompt}
The system prompt used for determining whether or not to leave the Engaged State and enter the Disengaged State.

\begin{lstlisting}
You should try to ask questions that would keep the conversation going, don't try to disengage unless the person has shown clear intent to disengage with you. 
Remember that you are having a hallway conversation. So if the conversation turns are over 5 be mindful of the user's time. 
Ask if they can stay longer to talk. You should always ask if they can stay longer to talk. 
Do not disengage until the person has responded to your disengagement request.

\end{lstlisting}

\section{Background Information and Social Relationship Information}

The UserContext that the model uses to infer knowledge about the Source and its social relationships.
The text colored in blue is the "Background Information" and "Social Relationship Information" that is updated each day across the study.

\begin{lstlisting}
{
  "$schema": "./userContext-schema.json",
  "Background": "You are [Source]'s IRL Ditto. You are a digital embodiment of [Source] on a wall. (...) Also tomorrow is your departing lunch wondering where you should go to lunch with your mentors- [names redacted]" ,
  "PersonalityTraits": "Cheerful, Sarcastic, Humorous, Witty",
  "SocialRelationshipInfo": [
    {
      "Who": "X",
      "RelationshipInfo": "A [role redacted] at [organization redacted]. He is working on a cool ring Project. X likes to go hiking. ",
      "SourceIntent": "Today you should talk about these topics only --  You are wondering about X's thoughts on the IRL Ditto, what he would like to see more from the Ditto. "
    },
    {
      "Who": "Y",
      "RelationshipInfo": "A [role redacted] at [organization redacted]. You helped Y with her research project by creating a cool dog park simulation for her. You're curious about her vision for the project and her honest thoughts on paddleboarding. Focus more on the topic for today",
      "SourceIntent": "Today you should talk about these topics only -- Thank Y for coming around each day to talk to you. Wish her good luck on her CHI paper. Don't keep her too long since she might be a bit busy today. If she has time, ask her about her thoughts on the IRL Ditto."
    }  
  ]
}
\end{lstlisting}

\section{Study Materials}

The IRL Ditto Zone included signage to notify individuals of active recording.

\begin{figure*}[!h]
    \centering 
    \includegraphics[width=1.0\textwidth]{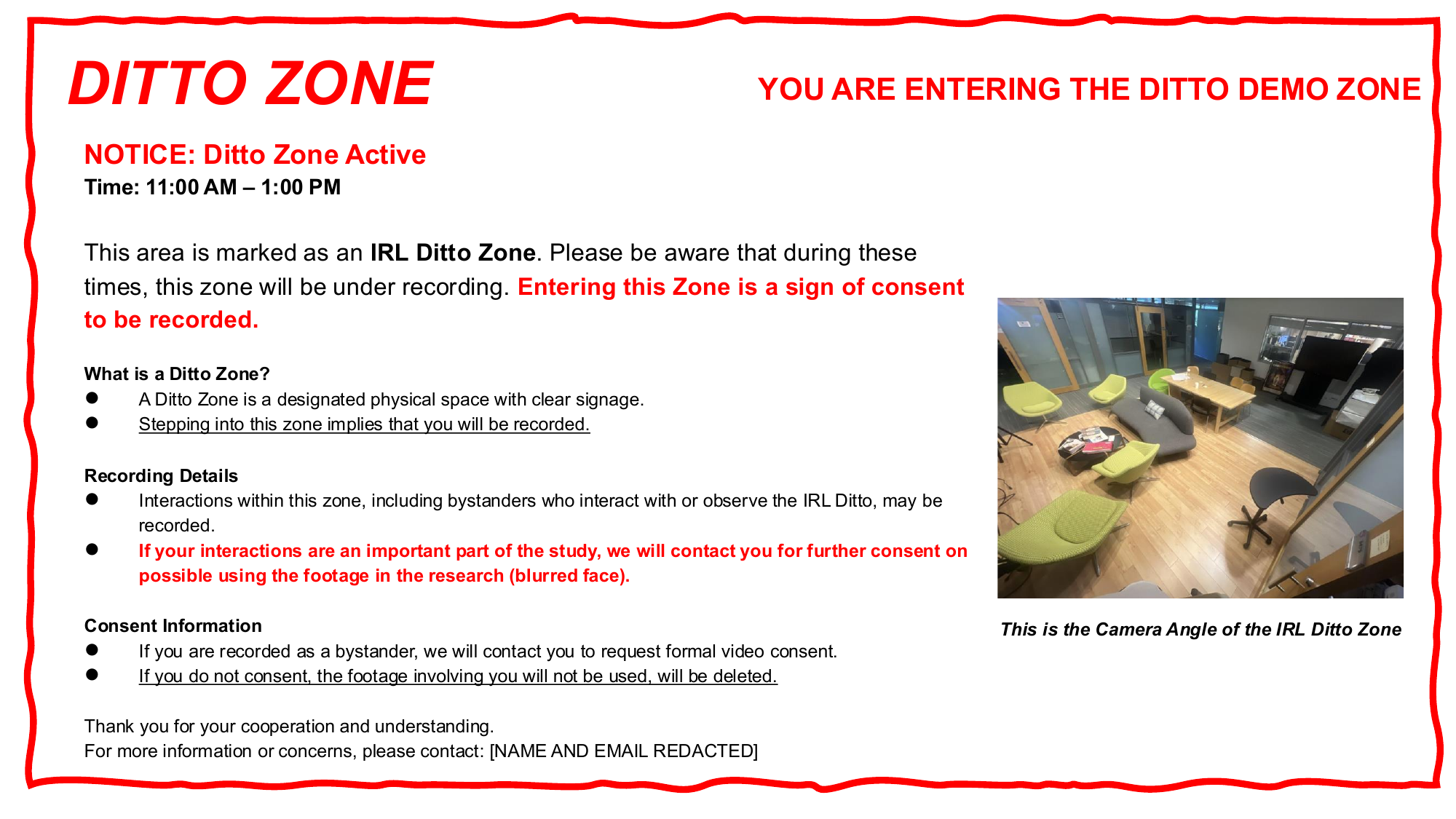}
    \caption[Image of Notice place in IRL Ditto Zone]{Signage indicating the active IRL Ditto Zone.}
    \label{fig:study-signage}
\end{figure*}

\end{document}